\documentclass{article}

\usepackage{arxiv}

\usepackage[utf8]{inputenc} % allow utf-8 input
\usepackage[T1]{fontenc}    % use 8-bit T1 fonts
\usepackage[pagebackref=true,breaklinks=true,colorlinks,bookmarks=false]{hyperref}
\usepackage{url}            % simple URL typesetting
\usepackage{booktabs}       % professional-quality tables
\usepackage{amsfonts}       % blackboard math symbols
\usepackage{nicefrac}       % compact symbols for 1/2, etc.
\usepackage{microtype}      % microtypography
\usepackage{graphics}
\usepackage{times}
\usepackage{epsfig}
\usepackage{amsmath}
\usepackage{amssymb}
\usepackage{booktabs,siunitx}
\usepackage{xcolor}
\usepackage{comment}
\usepackage{scrextend}
\usepackage{chngcntr}

\definecolor{my_color}{rgb}{0,0,0}
\definecolor{my_color_1}{rgb}{0,0,0}

\begin{document}

\title{Learning Sparse \& Ternary Neural Networks\\
with Entropy-Constrained \\Trained Ternarization (EC2T)}

\newcommand{\addr}{Department of Video Coding \& Analytics, Fraunhofer Heinrich Hertz Institute, Berlin, Germany}

\newcommand{\mails}{\footnotesize{\texttt{\{arturo.marbangonzalez, daniel.becking, simon.wiedemann, wojciech.samek\} @hhi.fraunhofer.de}}}

\author{
 Arturo Marban \\
   \And
 Daniel Becking \\
  \And
 Simon Wiedemann \\
  \And
 Wojciech Samek \\
}

\maketitle

%%%%%%%%% ABSTRACT
\begin{abstract}
Deep neural networks (DNN) have shown remarkable success in a variety of machine learning applications. The capacity of these models (i.e., number of parameters), endows them with expressive power and allows them to reach the desired performance. In recent years, there is an increasing interest in deploying DNNs to resource-constrained devices (i.e., mobile devices) with limited energy, memory, and computational budget. To address this problem, we propose Entropy-Constrained Trained Ternarization (EC2T), a general framework to create sparse and ternary neural networks which are efficient in terms of storage (e.g., at most two binary-masks and two full-precision values are required to save a weight matrix) and computation (e.g., MAC operations are reduced to a few accumulations plus two multiplications). This approach consists of two steps. First, a super-network is created by scaling the dimensions of a pre-trained model (i.e., its width and depth). Subsequently, this super-network is simultaneously pruned (using an entropy constraint) and quantized (that is, ternary values are assigned layer-wise) in a training process, resulting in a sparse and ternary network representation. We validate the proposed approach in CIFAR-10, CIFAR-100, and ImageNet datasets, showing its effectiveness in image classification tasks.
\end{abstract}

%%%%%%%%% BODY TEXT %%%%%%%%%
% Introduction.
\section{Introduction}
\label{section:introduction}

Convolutional neural networks (CNN) have excelled in numerous computer vision applications. Their performance is attributed to their design. That is, deeper (i.e., designed with many layers) and high-capacity (i.e., equipped with many parameters) CNNs achieve better performance in a given task, at the cost of sacrificing computational and memory efficiency.  This general trend has been disrupted by the need to deploy neural networks in resource-constrained devices (e.g., autonomous vehicles, robots, smartphones, wearable, and IoT devices) with limited energy, memory, and computational budget, as well as low-latency and/or low-communication cost requirements. Thus, driven by both the industry and the scientific community, the design of efficient CNNs has become an active area of research. Moreover, the Moving Picture Expert Group (MPEG) of the International Organization of Standards (ISO) joined this endeavor, and recently issued a call on neural network compression techniques~\cite{MPEG_Requirements:2019}.

Recent studies have shown that most CNNs are over-parameterized for the given task~\cite{PredictDnnParameters:M_Denil_2013}. Such models can be interpreted as super-networks,
designed with millions of parameters to reach a target performance (e.g., high classification accuracy), while being memory and computational inefficient. However, from these models, it is possible to find a small and efficient sub-network with comparable performance. This hypothesis has been validated with simple methods, i.e., by pruning neural network connections based on the weights' magnitude~\cite{Pruning:S_Han_2015}, resulting in little accuracy degradation. Moreover, the recently proposed lottery-ticket hypothesis~\cite{LotteryTicketHypothesis:J_Frankle_2018}, supports the existence of an optimal sub-network inside a super-network, and has shown to generalize across different datasets and optimizers~\cite{GenLotteryTicketHypothesis:Morcos_2019}.

Among existing network compression techniques, pruning and quantization are two popular and effective techniques to reduce the redundancy of deep neural networks~\cite{NNCompressionSurvey:Y_Cheng_2018}. Pruning entails systematically removing network connections in a structured (i.e., by removing groups of parameters) or unstructured fashion (i.e., by removing individual parameter elements)~\cite{StateNNPruning:D_Blalock_2020}. In contrast, quantization minimizes the bit-width of the network parameter values (and thus, the number of distinct values)~\cite{NetworkQuantization:Y_Choi_2016, TrainedTernaryQ:C_Zhu_2017}. From another perspective, efficient neural networks can be designed by finding the right balance between its dimensions, i.e., the networks' width, depth, and input resolution. In this regard, compound model scaling~\cite{EfficientNet:M_Tan_2019b} allows scaling the dimensions of a baseline-network according to some heuristic rules grounded on computational efficiency.

\begin{figure*}[!th]
	% Caption and label go in the first argument and the figure contents
	% go in the second argument
	\begin{center}
		\includegraphics[width=0.825\linewidth]{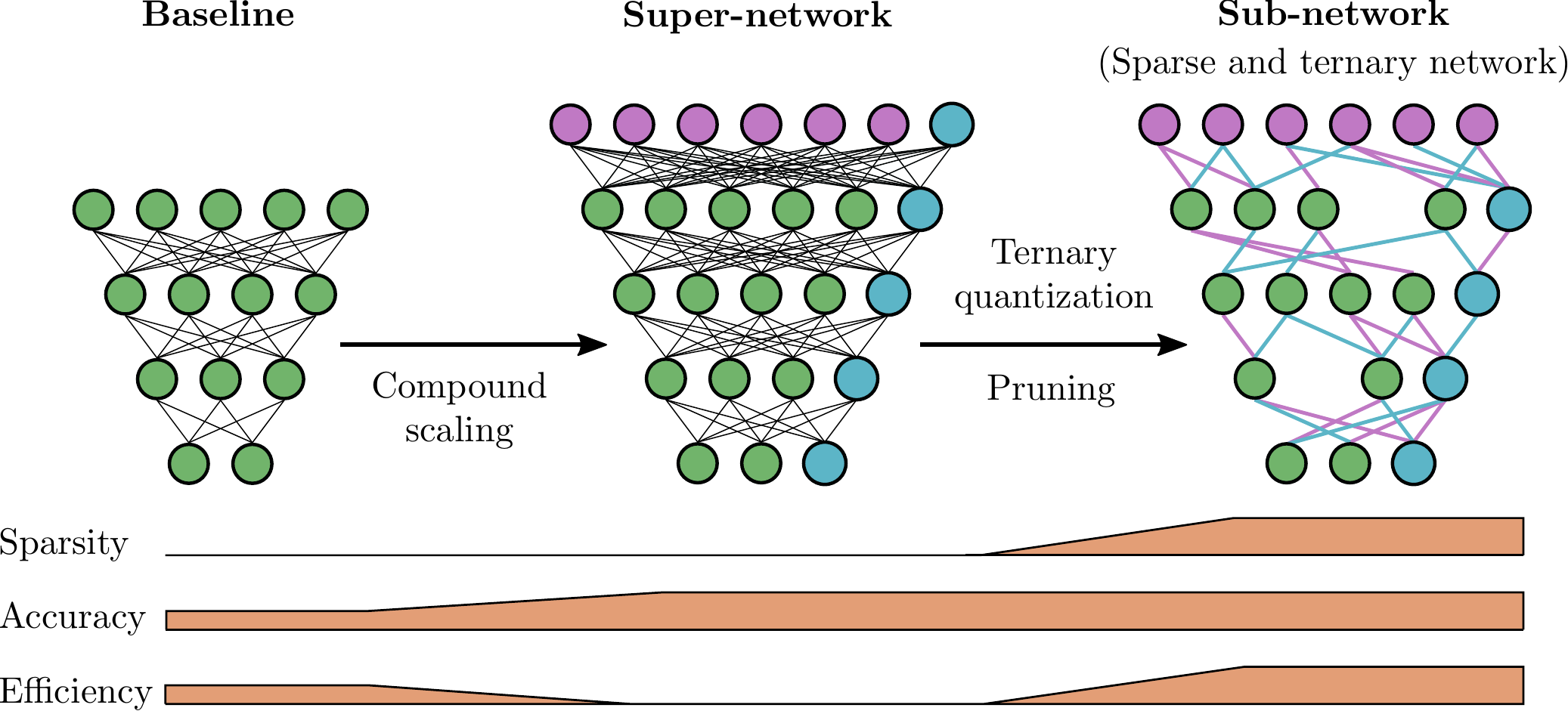}
	\end{center}
	\caption{In the EC2T approach, model compound scaling is used to create a super-network from a baseline-network. Afterward, in a ternary quantization stage, this super-network is simultaneously pruned and quantized, rendering a sparse and ternary sub-network with comparable performance.}
	\label{fig:proposed-method}
\end{figure*}

In this work, we propose Entropy-Constrained Trained Ternarization (EC2T), a method that leverages on compound model scaling~\cite{EfficientNet:M_Tan_2019b} and ternary quantization techniques~\cite{TrainedTernaryQ:C_Zhu_2017}, to design a sparse and ternary neural network. The motivations behind such network representation are based on efficiency. Specifically, in terms of storage, at most two binary-masks and two full-precision values are required to represent and save each layer's weight matrix. Regarding mathematical operations, multiply-accumulate operations (MACs) are reduced to a few accumulations plus two multiplications. The EC2T approach is illustrated in~Figure~\ref{fig:proposed-method} and consists of two stages. In the first stage, a super-network is created by scaling the dimensions of a baseline-network (its width and depth). Subsequently, during a training stage, a sparse and ternary sub-network is found by simultaneously pruning (enforced by introducing an entropy constraint in the assignment cost function) and quantizing (ternary values are assigned layer-wise) the super-network. Specifically, our contributions are:

\begin{itemize}
	\item We propose an approach to design sparse and ternary neural networks, that relies on compound model scaling~\cite{EfficientNet:M_Tan_2019b} and quantization techniques. For the latter, we extend the approach described in~\cite{TrainedTernaryQ:C_Zhu_2017} by introducing an assignment cost function in terms of distance and entropy constraints.  The entropy constraint allows adjusting the trade-off between sparsity and accuracy in the quantized model. Therefore, quantized models with different levels of sparsity can be rendered, according to the compression and application requirements. 
	\item Our approach allows simultaneous quantization and sparsification in a single training stage.
	\item In the context of image classification, the proposed approach finds sparse and ternary networks across different datasets (CIFAR-10, CIFAR-100, and ImageNet), whose performance is competitive with efficient state-of-the-art models.
\end{itemize}

This paper is organized as follows. First, in section~\ref{section:related_works}, a literature review of techniques to design efficient neural networks is provided, emphasizing those that are related to our approach. Subsequently, in section~\ref{section:method}, the proposed EC2T approach is detailed. Afterward, in section~\ref{section:experiments_and_results}, we present experimental evidence and results, validating the proposed method across different networks and datasets. Finally, in section~\ref{section:conclusions}, we discuss the insights of the EC2T approach, its advantages and downsides, and future work.
% Related Works.
\section{Related Works}
\label{section:related_works}
In recent years, various techniques have been proposed in the literature to design efficient neural networks, e.g., pruning, quantization, distillation, and low-rank factorization~\cite{NNCompressionSurvey:Y_Cheng_2018}. In particular, pruning and quantization provide unique benefits to DNNs in terms of hardware efficiency and acceleration.

Pruning removes non-essential neural network connections, according to different criteria, either in groups (structured pruning) or individual parameters (unstructured pruning). Specifically, the second approach is achieved by maximizing the sparsity~\footnote[1]{Percentage of zero-valued parameter elements in the whole neural network.} of the network parameters. \textcolor{my_color_1}{Consequently, the computational complexity of the network is reduced, since arithmetic operations can be skipped for those parameter elements which are zero~\cite{EfficientRepDNN:S_Wiedemann_2020}}. Early works on sparsity use second-order derivatives (Hessian) to compute the saliency of parameters, suppressing those with the smallest value~\cite{PruningWithHessian:Y_LeCun_1990, PruningWithHessian:Hassibi_1993}. Current state-of-the-art techniques to promote sparsity in DNNs rely either on magnitude-based pruning or Bayesian approaches~\cite{StateSparsityDNNs:Gale_2019}. Magnitude-based pruning is the simplest and most effective way to induce sparsity in neural networks,~\cite{StateNNPruning:D_Blalock_2020}. In contrast, Bayesian approaches although computationally expensive, represent an elegant solution to the problem. Moreover, they establish connections with information theory. In this context, variational dropout~\cite{VarDropout:D_Molchanov_2017} and $l_0$-regularization~\cite{SparseNetworksL0Reg:C_Louizos_2017b} are two representative techniques.

Regarding~quantization, it reduces the redundancy of deep neural networks by minimizing the bit-width of the full-precision parameters. Therefore, quantized networks require fewer bits to represent each full-precision weight, and demand less mathematical operations than their full-precision counterparts. Binary networks~\cite{BinarizedNN:M_Courbariaux_2016, BNNSurvey:T_Simons_2019} represent an extreme case of quantization where both, weights and activations are binarized. Thus, arithmetic operations are reduced to bit-wise operations. By introducing three distinct elements per layer, ternary networks achieve more expressive power and higher performance than binary networks. Moreover, sparsity can be induced in the network by including zero as a quantized value, while the remaining values are modeled with scaling factors per layer. Following this approach,~\cite{TernaryWeightNetworks:F_Li_2016} proposed to  minimize the Euclidean distance between full-precision and quantized parameters (e.g., $w_q$), where the latter are symmetrically constrained (e.g., $w_q\in\{-a,0,a\}$, with $a>0$). In contrast,~\cite{TrainedTernaryQ:C_Zhu_2017} used asymmetric constraints (e.g., $w_q\in\{-a,0,b\}$, with $a>0$ and $b>0$), improving the modeling capabilities of ternary networks. Several variants of ternary network quantization exist, e.g., based on Truncated Gaussian Approximation (TGA)~\cite{TernaryNetworks_TGA:Z_He_2019}, Alternating Direction Method of Multipliers (ADMM))~\cite{TernaryNetworks_ADMM:Leng_2018}, and Multiple-Level-Quantization~(MLQ)~\cite{TernaryNetworks_MLQ:Y_Xu_2018}, among others. With regards to hardware efficiency, ternary networks represent a trade-off between binary networks (extremely hardware-friendly, but with limited modeling capabilities) and their full-precision counterparts (with higher modeling capabilities, but expensive in terms of storage and computational resources),~\cite{TernaryWeightNetworks:F_Li_2016}.

Usually, highly efficient network representations are the result of combining multiple techniques. For instance, pruning followed by quantization ~\cite{BayesianCompression:M_Federici_2017, EntropyConstTraining:S_Wiedemann_2019}, in addition to entropy coding~\cite{DeepCompression:S_Han_2016, BayesianCompression:C_Louizos_2017a, DeepCABAC:Journal:S_Wiedemann_2020}. From a different perspective, progress in designing efficient neural networks has been fueled by advances in hand-crafted architectures (e.g., Mobilenet~\cite{MobileNet:A_Howard_2017}, Mobilenet-V2~\cite{MobileNetV2:M_Sandler_2018}, and ShuffleNet~\cite{ShuffleNet:X_Zhang_2018}) as well as neural architecture search techniques (e.g., Mnasnet~\cite{MnasNet:M_Tan_2019a}, EfficientNet~\cite{EfficientNet:M_Tan_2019b}, and MobileNet-V3~\cite{MobileNetV3:A_Howard_2019}). Moreover, simpler methods such as model scaling, allows increasing the performance of a baseline network by scaling one or more dimensions (i.e., its depth, width, and input resolution) independently~\cite{MnasNet:M_Tan_2019a, MobileNetV3:A_Howard_2019}. In~\cite{EfficientNet:M_Tan_2019b}, this approach is improved with the introduction of compound model scaling, where the network dimensions are treated as dependent variables, constrained by a limited number of resources, measured in terms of floating-point operations (FLOPs).

In this research work, we advocate for compound model scaling, ternary quantization, and information theory techniques, as the core building blocks to design a CNN with optimal dimensions (i.e., the right balance between the networks' width and depth) and efficient parameter representation (i.e., three distinct values per layer and maximal sparsity).
% and models can be efficiently stored or transmitted over communication channels, e.g., by applying entropy coding techniques~\cite{DeepCABAC:Journal:S_Wiedemann_2020}
% Method.
\section{Learning Sparse \& Ternary Networks}
\label{section:method}
The entropy-constrained trained ternarization (EC2T) approach (see~Figure~\ref{fig:proposed-method}), consists of two stages, namely compound model scaling followed by ternary quantization, both described in sections~\ref{section:method:compound_model_scaling} and~\ref{section:method:ternary_quantization}, respectively.

\subsection{Compound model scaling}
\label{section:method:compound_model_scaling}
In this stage, a super-network is created by scaling the dimensions of a pre-trained model, resulting in an over-parameterized network. Specifically, the pre-trained network's depth, width, and input image resolution, are modified with the scaling factors $d$, $w$, and $r$, respectively, according to~Equation~(\ref{eq:compound_model_scaling}). In this equation, $a$, $b$ and $c$, are constants determined by grid search, and $\phi$ is an user specified parameter. For small-scale datasets (CIFAR-10 and CIFAR-100) the input image resolution was fixed in the pre-trained model. Thus,~Equation~(\ref{eq:compound_model_scaling}) was solved with $r=1$. On the other hand, for large-scale datasets (ImageNet), the EfficientNet-B1 network was adopted using the scaling factors suggested in~\cite{EfficientNet:M_Tan_2019b}.
\begin{align}\label{eq:compound_model_scaling}
&d=a^{\phi},~w=b^{\phi},~r=c^{\phi} \\
&\quad\quad\text{s.t.}~~~a \cdot b^2 \cdot c^2 \approx 2~~~\text{and}~~~a \geq 1, b \geq 1, c \geq 1 \nonumber
\end{align}
\subsection{Ternary quantization}
\label{section:method:ternary_quantization}
\begin{figure*}[!th]
	% Caption and label go in the first argument and the figure contents
	% go in the second argument
	\begin{center}
		\includegraphics[width=0.85\linewidth]{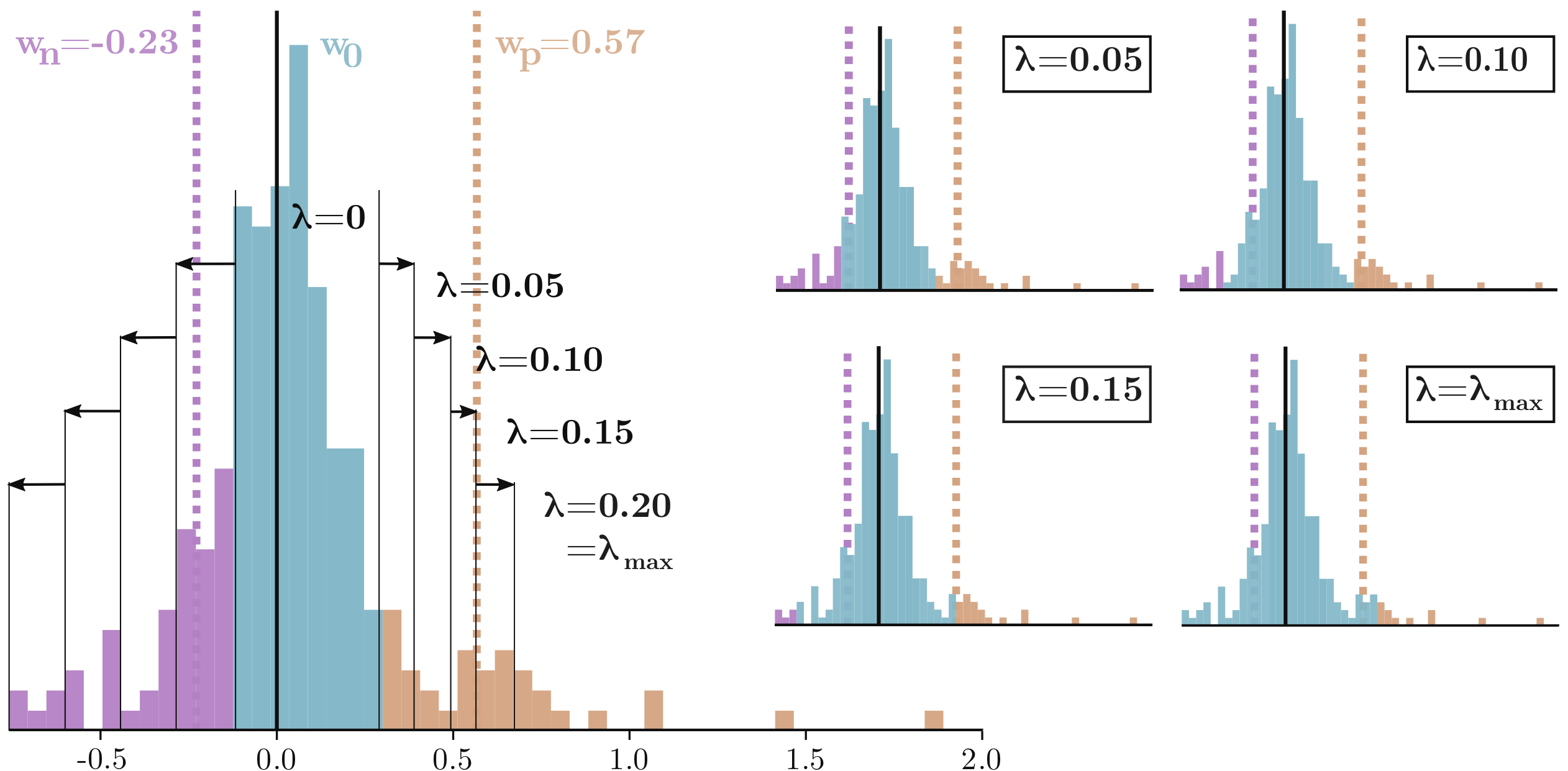}
	\end{center}
	\caption{Histograms of the parameters in the projection-convolution layer, in the first block (MBConv1) of the EfficientNet-B1 network. The centroid values $w_n$ (negative scalar), $w_0$ (zero), and $w_p$ (positive scalar), are shown in magenta, blue and orange colors, respectively. The hyper-parameter $\lambda$ controls the intensity of the network sparsification, i.e., how many full-precision weight elements are assigned with the value $w_0$. When $\lambda$=0, the weights are quantized to their nearest neighbor centroids. Using small values for $\lambda$ (see the histogram with $\lambda$=0.05) results in quantized parameters with low sparsity (i.e., few parameters are set to zero). As $\lambda$ is increased (see histograms with $\lambda$=0.10 and 0.15), the sparsity of the quantized parameters is promoted (i.e., most parameters are set to zero). Eventually, as this process continues, there is a value $\lambda=\lambda_{max}$, at which the network parameters are binarized. In this special case, one of the two clusters of values (represented by $w_n$ and $w_p$) is completely assigned to $w_0$ (see the histogram with $\lambda$=$\lambda_{max}$).}
	\label{fig:ternary-assigment}
\end{figure*}
In this stage, a sparse and ternary sub-network is obtained by simultaneously pruning and quantizing a super-network. To this end, we extend the approach described in ~\cite{TrainedTernaryQ:C_Zhu_2017}, where a ternary network is obtained by the inter-play between quantized and full-precision models. That is, gradients from the quantized model are used to update both, its parameters and those of the full-precision model. Therefore, the first parameter update enables the learning of ternary values \textcolor{my_color_1}{(i.e., only two scalar values per layer are learned, while the third quantized value, which is zero, is excluded from the learning process)}. On the other hand, the latter parameter update promotes the learning of ternary assignments (i.e., by adapting the full-precision parameters to the quantization process). Nonetheless, this approach does not allow explicit control of the sparsification process. To overcome this limitation, we introduce the assignment cost function shown in~Equation~(\ref{eq:assignment_cost_function}), which guides the assignment (with centroid indices) of ternary values (or centroid values) in the quantized network, in terms of distance and entropy constraints. 
\begin{equation}\label{eq:assignment_cost_function}
\textbf{C}_c^{(l)} = d(\textbf{W}^{(l)}, w_c^{(l)}) - \lambda^{(l)}~\log_2(P_c^{(l)})
\end{equation}
\begin{equation}\label{eq:distance_function}
d_{W_{ij},w_c}= (W_{ij} - w_c)^2
\end{equation}
In~Equation~(\ref{eq:assignment_cost_function}), $\textbf{C}_c^{(l)}$ stands for the assignment cost for the full-precision weights $\textbf{W}^{(l)}$ at layer $l$, given the centroid values $w_c^{(l)}$, indexed by $c$. Therefore, if $\textbf{W}^{(l)}$ has $m\times n$ dimensions and there are $n_{c}$ centroid values in that layer, then $\textbf{C}_c^{(l)}\in\Re^{n_{c}\times m\times n}$. The first term in~Equation~(\ref{eq:assignment_cost_function}) measures the distance between every full-precision weight element $W_{ij}^{(l)}\in\textbf{W}^{(l)}$ (where $i$ and$j$ are indices along the dimensions of $\textbf{W}^{(l)}$) and the centroid values $w_c^{(l)}\in\Re$, according to~Equation~(\ref{eq:distance_function}). The second term in~Equation~(\ref{eq:assignment_cost_function}), weighted by the scalar $\lambda^{(l)}\in\Re$, is an entropy constraint which promotes sparsity in the quantized model. This is achieved by measuring the information content of the quantized weights, i.e., $I=-\log_2(P_c^{(l)})\in\Re$, where the probability $P_c^{(l)}\in[0,1]$ defines how likely a weight element $W_{ij}^{(l)}\in\textbf{W}^{(l)}$ is going to be assigned to the centroid value $w_c^{(l)}$. This probability is calculated for each layer $l$ as $P_c^{(l)} = N_{w_c}^{(l)} / N_{\textbf{W}}^{(l)}$, with $N_{w_c}^{(l)}$ being the number of full-precision weight elements assigned to the centroid value $w_c^{(l)}$, and $N_{\textbf{W}}^{(l)}$ the total number of parameters in $\textbf{W}^{(l)}$.

After computing~Equation~(\ref{eq:assignment_cost_function}) (for all layers and centroid values), the quantized model is updated at layer $l$, by assigning the current centroid values ($w_c^{(l)}$), using the new centroid indices ($c$) obtained from~Equation~(\ref{eq:assignment}). In this equation, the assignment matrix $\textbf{A}^{(l)}$ has the  dimensions of the full-precision weights $\textbf{W}^{(l)}$. For ternary networks, we define the centroid values as $w_c^{(l)} \in \{w_n, w_0, w_p\}$, and their assignments with the indices $c \in \{n, 0, p\}$. In this notation, the indices $n$, $0$, and $p$, correspond to negative, zero, and positive values, respectively. 
\begin{equation}\label{eq:assignment}
\textbf{A}^{(l)} = \underset{c}{\text{argmin}} \ \textbf{C}_c^{(l)}
\end{equation}
\begin{figure*}[!th]
	\begin{center}
		\includegraphics[width=\linewidth]{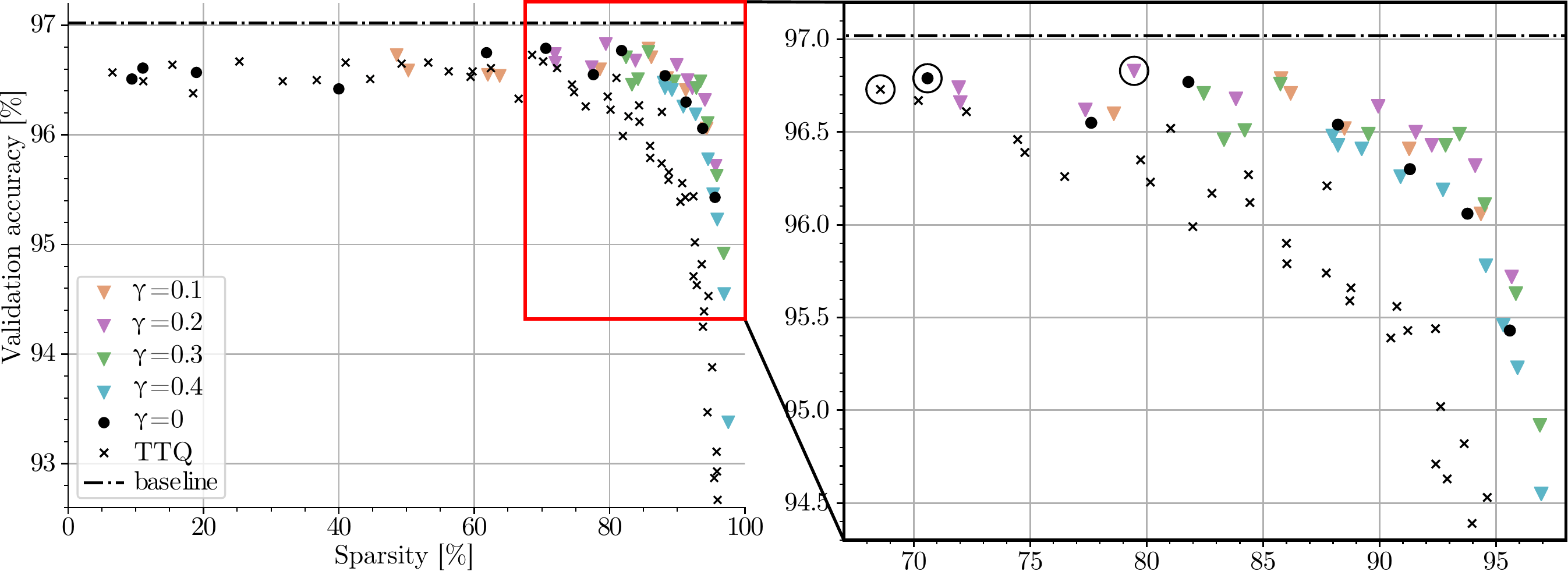}
	\end{center}
	\caption{Performance of the C10-MicroNet network evaluated in the CIFAR-10 dataset, using TTQ vs our proposal (EC2T). Every data point in this plot represents a quantized model, trained with a specific level of sparsity, and initialized with different centroid values. In the TTQ approach, the sparsity is controlled via simple thresholding as described in~\cite{TrainedTernaryQ:C_Zhu_2017}, whereas in the EC2T approach, it is modulated by $\gamma$, which was increased from 0.0 (low sparsity) to 0.4 (high sparsity), in steps of 0.1. Notice that beyond 70\% sparsity, the accuracy of the quantized models degrades quickly. However, this effect is more evident when using TTQ than EC2T.}
	\label{fig:c10-micronet-cifar10:ttq-vs-ec2t}
\end{figure*}
During the ternary quantization process, the strength of the sparsification (at layer $l$) is modulated by the scalar $\lambda^{(l)}$ (shown~in~Equation~(\ref{eq:assignment_cost_function})). As a concrete example,~Figure~\ref{fig:ternary-assigment} illustrates the effect of using different values for $\lambda^{(l)}$ during the quantization of the parameters (in the first block) of the EfficientNet-B1 network. In practice, $\lambda^{(l)}$ is computed as $\lambda^{(l)} = \gamma~\delta^{(l)}~\lambda_{\max}^{(l)}$. In this expression, $\gamma$ is a global hyper-parameter that controls the intensity of the sparsification, while $\delta^{(l)}$ and $\lambda_{\max}^{(l)}$ are scalars computed layer-wise.~The scaling factor $\delta^{(l)}$, renders higher values for layers with lots of parameters. Analogously, it renders lower values for layers with few parameters. Finally, $\lambda_{\max}$ is updated during training and avoids a binary quantization process (see the histogram with $\lambda$=$\lambda_{max}$ in~Figure~\ref{fig:ternary-assigment}).
% Experiments \& Results.
\section{Experiments \& Results}
\label{section:experiments_and_results}
% Table: EC2T vs TTQ.
\begin{table*}[!th]
\renewcommand{\arraystretch}{1.0}
\caption{Comparison of the EC2T approach vs state-of-the-art ternary network quantization techniques, applied to ResNet-20 and ResNet-18 networks, in CIFAR-10 and ImageNet datasets, respectively.}
\label{table:ec2t_vs_other}	
\begin{center}
	\begin{tabular}{lrrrrrr}
		\toprule
		\bfseries Model & \bfseries Top-1 Acc.~($\%$) & \bfseries $\frac{|W=0|}{|W|}$~($\%$)$^\ddagger$ & \bfseries $\#$Params.  & \bfseries $\#+$ & \bfseries $\#\times$ & \bfseries $\#$FLOPs \\
		\midrule
		\multicolumn{7}{c}{\bfseries ImageNet} \\
		\midrule
		\bfseries ResNet-18$^{a}$  & 69.75 & 0.00 & 11M & 1795M & 1797M & 3592M \\	
		EC2T-1 ($\lambda=0$)$^{b}$ & 67.30 & 26.80 & 852K & 669M & 59M & 728M\\
		EC2T-2 ($\lambda>0$)$^{c}$ & \bfseries67.58 & 59.00 & 734K & 560M & 61M & 622M\\
		EC2T-3 ($\lambda>0$)$^{c}$ & 67.26 & 72.09 & 686K & 528M & 57M & 585M\\
		EC2T-4 ($\lambda>0$)$^{c}$ & 67.02 & \bfseries~75.62 & \bfseries~673K & \bfseries~424M & \bfseries~57M & \bfseries~481M\\
		TTQ~\cite{TrainedTernaryQ:C_Zhu_2017} 	   	& 66.60  & 30-50 & {\scriptsize $\oslash$} & {\scriptsize $\oslash$} & {\scriptsize $\oslash$} & {\scriptsize $\oslash$} \\
		ADMM~\cite{TernaryNetworks_ADMM:Leng_2018} 	& 67.00 & {\scriptsize $\oslash$} & {\scriptsize $\oslash$} & {\scriptsize $\oslash$} & {\scriptsize $\oslash$} & {\scriptsize $\oslash$} \\
		TGA~\cite{TernaryNetworks_TGA:Z_He_2019}   	& 66.00 & {\scriptsize $\oslash$} & {\scriptsize $\oslash$} & {\scriptsize $\oslash$} & {\scriptsize $\oslash$} & {\scriptsize $\oslash$} \\
		\midrule
		\multicolumn{7}{c}{\bfseries CIFAR-10} \\
		\midrule
		\bfseries ResNet-20$^{a}$ & 91.67 & 0.00 & 269K & 40.6M & 40.7M & 81.3M\\
		EC2T-1 ($\lambda=0$)$^{b}$ & \bfseries~91.16 & 45.17 & 13.4K & 10.6M & 0.5M & 11.1M \\
		EC2T-2 ($\lambda>0$)$^{c}$ & 91.01 & 63.90 & 11.8K & 8.0M & 0.5M & 8.5M \\
		EC2T-3 ($\lambda>0$)$^{c}$ & 90.76 & \bfseries~73.26 & \bfseries~11.0K & \bfseries~6.1M & \bfseries~0.5M & \bfseries~6.6M \\
		TTQ~\cite{TrainedTernaryQ:C_Zhu_2017} & 91.13 & 30-50 & {\scriptsize $\oslash$} & {\scriptsize $\oslash$} & {\scriptsize $\oslash$} & {\scriptsize $\oslash$} \\
		TGA~\cite{TernaryNetworks_TGA:Z_He_2019} & 90.39 & {\scriptsize $\oslash$} & {\scriptsize $\oslash$} & {\scriptsize $\oslash$} & {\scriptsize $\oslash$} & {\scriptsize $\oslash$} \\
		MLQ~\cite{TernaryNetworks_MLQ:Y_Xu_2018} & 90.02 & {\scriptsize $\oslash$} & {\scriptsize $\oslash$} & {\scriptsize $\oslash$} & {\scriptsize $\oslash$} & {\scriptsize $\oslash$} \\
		\bottomrule
	\end{tabular}\\
	\footnotesize{$^{a}$ Baseline model. $^{b}$ EC2T approach with the entropy constraint disabled ($\lambda=0$).} \\
	\footnotesize{$^{c}$ EC2T approach with the entropy constraint enabled ($\lambda>0$).} \\
	\footnotesize{$^\ddagger$ Sparsity, measured as the percentage of zero-valued parameters in the whole neural network.} \\
	\footnotesize{{\scriptsize $\oslash$}: Not reported by the authors.}
\end{center}
\end{table*}
The experiments were conducted in a variety of networks across different datasets (i.e., CIFAR-10, CIFAR-100, and ImageNet), using multiple GPUs (NVIDIA Titan-V and Tesla-V100).

First, to reveal the advantages of our proposal (EC2T) over Trained-Ternary-Quantization (TTQ)~\cite{TrainedTernaryQ:C_Zhu_2017}, an image classification network was designed for the CIFAR-10 dataset, by introducing the building blocks of PyramidNet~\cite{DeepPyramidalResnets:D_Han_2017} in the ResNet-44 architecture~\cite{Resnet:K_He_2016}. This neural network, termed C10-MicroNet, was derived from models designed for the 2019~MicroNet Challenge~\footnote[2]{\label{micronet-website}\url{https://micronet-challenge.github.io}} competition. \textcolor{my_color_1}{For a detailed description of the network architecture, see Appendix~\ref{appendix:micronet-c10-c100-networks}}. The experimental results contrasting the two mentioned approaches are depicted in~Figure~\ref{fig:c10-micronet-cifar10:ttq-vs-ec2t}. In this illustration, notice that as the sparsity of the quantized networks increases, EC2T shows less accuracy degradation than TTQ.

Subsequently, Table~\ref{table:ec2t_vs_other} provides a comparison of the EC2T approach vs state-of-the-art ternary quantization techniques, by applying them to ResNet-20 and ResNet-18 networks, in CIFAR-10 and ImageNet datasets, respectively. From these results, we have two main conclusions. First, they suggest that disabling the entropy constraint in Equation~(\ref{eq:assignment_cost_function}) (i.e., setting $\lambda=0$), renders ternary models with low sparsity. Nonetheless, they are more efficient than their full-precision counterparts and show little accuracy degradation. These ternary networks are referred to as EC2T-1 in Table~\ref{table:ec2t_vs_other}. Specifically, in the ImageNet dataset, the EC2T-1 model reduces the parameter count in 92.25\% and the FLOPs in 79.73\%, while in the CIFAR-10 dataset, the reductions are 95.02\% and 86.35\% in parameter count and FLOPs, respectively. \textcolor{my_color_1}{In contrast, by enabling the entropy constraint in Equation~(\ref{eq:assignment_cost_function}) (i.e., setting $\lambda>0$), it results in ternary models with increased sparsity, and thus, they are more efficient in terms of parameter size and mathematical operations. For instance, in the ImageNet dataset, the model with the highest sparsity is EC2T-4, which reduces the number of parameters by 93.88\% and the number of FLOPs by 86.61\%, while its accuracy is degraded only by 2.73\%. Likewise, in the CIFAR-10 dataset, the model with the highest sparsity is EC2T-3, with an accuracy degradation of 0.91\%, while the parameter count and FLOPs are reduced by 95.91\% and 91.88\%, respectively. The second conclusion is that the EC2T approach renders accurate ternary models, which are competitive with state-of-the-art techniques. Regarding sparsity, only~\cite{TrainedTernaryQ:C_Zhu_2017} provides an estimated value for the ternary models after applying TTQ (30\%-50\%). For the remaining techniques (ADMM~\cite{TernaryNetworks_ADMM:Leng_2018}, TGA~\cite{TernaryNetworks_TGA:Z_He_2019}, and MLQ~\cite{TernaryNetworks_MLQ:Y_Xu_2018}), only the quantized model accuracy is reported.}

Finally, Table~\ref{table:results_micronet} contrasts efficient state-of-the-art neural networks vs sparse and ternary networks rendered with our proposal, in three distinct datasets (CIFAR-10, CIFAR-100, and ImageNet). The former models include CondenseNet~\cite{Condensenet:G_Huang_2018}, Mobilenet-V2~\cite{MobileNetV2:M_Sandler_2018}, and Mobilenet-V3~\cite{MobileNetV3:A_Howard_2019}. The latter models result from applying the EC2T approach to the pre-trained networks, C10-MicroNet, C100-MicroNet, and EfficientNet-B1~\cite{EfficientNet:M_Tan_2019b}. In particular, the C10-MicroNet and C100-MicroNet networks were designed and improved based on our submissions to the 2019-MicroNet Challenge. \textcolor{my_color_1}{Both share the same topology, except in the last layer (i.e., the softmax layer), which is adapted to the number of output classes~(see Appendix~\ref{appendix:micronet-c10-c100-networks})}. From the results in Table~\ref{table:results_micronet}, we highlight two points. First, the ternary networks found by our proposed technique (see models indicated with EC2T), are more efficient in terms of parameter size and FLOPs than their respective baselines (C10-MicroNet, C100-MicroNet, and EfficientNet-B1). \textcolor{my_color_1}{Moreover, using the tree adder~\cite{StudyNetworkCompression:J_Cheng_2019} and efficient matrix representations (including Compressed-Entropy-Row (CER)/Compressed-Sparse-Row (CSR) formats~\cite{EfficientRepDNN:S_Wiedemann_2020} and the method described in Appendix~\ref{appendix:efficient-parameter-storage}), leads to further savings in mathematical operations and storage~(see models referred with Improvements).} Second, these ternary models are competitive with current state-of-the-art efficient neural networks~(i.e., CondenseNet, Mobilenet-V2, and Mobilenet-V3), offering similar advantages in terms of memory and computational resources.
% Table: Results of the MicroNet Challenge
\begin{table*}[!th]
\renewcommand{\arraystretch}{1.0}
\caption{Ternary models rendered with the EC2T approach vs efficient state-of-the-art neural networks, in CIFAR-10, CIFAR-100, and ImageNet datasets.}
\label{table:results_micronet}
\begin{center}
	\begin{tabular}{lrrrrrr}
		\toprule
		\bfseries Model & \bfseries Top-1 Acc. ($\%$) & \bfseries $\frac{|W=0|}{|W|}$~($\%$)$^\ddagger$ & \bfseries $\#$Params. & \bfseries $\#+$ & \bfseries $\#\times$ & \bfseries $\#$FLOPs \\
		\midrule
		\multicolumn{7}{c}{\bfseries ImageNet} \\
		\midrule
		\bfseries EfficientNet-B1$^{a}$ 			& 78.43 & 0.00 	& 7.72M 		& 654M 	& 670M 	& 1324M \\
		$+$EC2T~($\lambda>0$)$^{b}$ 				& 75.05 & 60.73 & 1.07M			& 338M 	& 50M 	& 387M \\
		\vspace{5pt}
		$\quad+$Improvements$^{c}$ 		& - 	& - 	& \textbf{972K} & \textbf{212M} 	& \textbf{50M} 	& \textbf{261M}\\
		MobileNet-V2 (d=1.4) 						& 74.70 & {\scriptsize $\oslash$} 	& 6.90M 		& {\scriptsize $\oslash$} 	& {\scriptsize $\oslash$} 	& 585M$^\star$ \\
		MobileNet-V3 (Large) 						& \textbf{75.20} & {\scriptsize $\oslash$} 	& 5.40M 		& {\scriptsize $\oslash$} 	& {\scriptsize $\oslash$} 	& 219M$^\star$ \\
		\midrule
		\multicolumn{7}{c}{\bfseries CIFAR-100} \\
		\midrule
		\bfseries C100-MicroNet$^{a}$ 				& 81.47 & 0.00 	& 8.03M 		& 1243M & 1243M & 2487M\\
		$+$EC2T~($\lambda>0$)$^{b}$ 				& 80.13 & 90.49 & 412K 			& 126M 	& 3M 	& 129M \\
		\vspace{5pt}
		$\quad+$Improvements$^{c}$ 		& - 	& - 	& \textbf{226K} & \textbf{67M} 	& \textbf{3M} 	& \textbf{71M} \\
		CondenseNet-86  							& 76.36 & {\scriptsize $\oslash$} 	& 520K 			& {\scriptsize $\oslash$} 	& {\scriptsize $\oslash$} 	& 65M$^\star$  \\							
		CondenseNet-182 							& \textbf{81.50} & {\scriptsize $\oslash$} 	& 4.20M 		& {\scriptsize $\oslash$} 	& {\scriptsize $\oslash$} 	& 513M$^\star$ \\
		\midrule
		\multicolumn{7}{c}{\bfseries CIFAR-10} \\
		\midrule
		\bfseries C10-MicroNet$^{a}$ 				& 97.02 & 0.00 	& 8.02M			& 1243M & 1243M & 2487M \\
		$+$EC2T~($\lambda>0$)$^{b}$ 				& 95.87 & 95.64 & 295K 			& 72M	& 3M 	&  75M  \\
		\vspace{5pt}
		$\quad+$Improvements$^{c}$			& - 	& - 	& \textbf{133K}	& \textbf{39M} 	& \textbf{3M} 	& \textbf{42M}  \\
		CondenseNet-86 								& 95.00 & {\scriptsize $\oslash$} 	& 520K 			& {\scriptsize $\oslash$} 	& {\scriptsize $\oslash$} 	& 65M$^\star$   \\
		CondenseNet-182  							& \textbf{96.24} & {\scriptsize $\oslash$} 	& 4.20M 		& {\scriptsize $\oslash$} 	& {\scriptsize $\oslash$} 	& 513M$^\star$  \\	
		\bottomrule
	\end{tabular} \\
	\footnotesize{$^{a}$ Baseline model. $^{b}$ EC2T approach with the entropy constraint enabled ($\lambda>0$).} \\
	\footnotesize{$^{c}$ Improved representation of the neural network parameters by applying the tree adder, \\
	the Compressed-Entropy-Row (CER)/Compressed-Sparse-Row (CSR) formats, and the method described Appendix~\ref{appendix:efficient-parameter-storage}.} \\
	\footnotesize{$^\ddagger$ Sparsity, measured as the percentage of zero-valued parameters in the whole neural network.} \\
	\footnotesize{$^\star$ Reported as Multiply-Additions (MAdds). The number of FLOPs is approximately twice this value.} \\
	\footnotesize{{\scriptsize $\oslash$}: Not reported by the authors.}
\end{center}
\end{table*}

% Conclusions.
\section{Conclusions}
\label{section:conclusions}

In this work, we presented Entropy-Constrained Trained Ternarization, an approach that relies on compound model scaling and ternary quantization to design efficient neural networks. By incorporating an entropy constraint during the network quantization process, a sparse and ternary model is rendered, which is efficient in terms of storage and mathematical operations. The proposed approach has shown to be effective in image classification tasks in both, small and large-scale datasets. As future work, this method will be investigated in other tasks and scenarios, e.g., federated-learning~\cite{FedLearning:F_Sattler_2019}. Moreover, interpretability techniques~\cite{LRP:S_Bach_2015} will help us to understand how these models make predictions given their constrained parameter space.
%%%%%%%%% Acknowledgement %%%%%%%%%
\section*{Acknowledgement}
This work was funded by the German Ministry for Education and Research as BIFOLD 
- Berlin Institute for the Foundations of Learning and Data (ref. 01IS18025A and ref 01IS18037I).
%%%%%%%%% Bibliography %%%%%%%%%
{\small
\bibliographystyle{unsrt}
\bibliography{EC2T_preprint.bbl}
}
\newpage
\appendix
\counterwithin{figure}{section}
\counterwithin{table}{section}
\onecolumn
\section{MicroNet-C10 \& MicroNet-C100 Networks}
\label{appendix:micronet-c10-c100-networks}
\begin{figure*}[h!]
	\centering
	\includegraphics[width=0.75\textwidth]{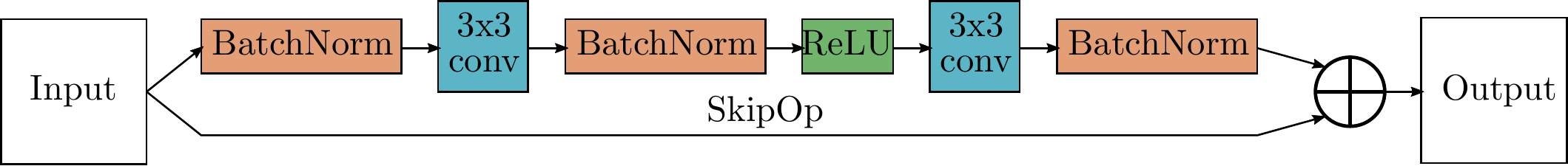}\\
	\caption{Building block for the baseline models, MicroNet-C10 and MicroNet-C100.}
	\label{fig:block}
\end{figure*}

The MicroNet-C10 and MicroNet-C100 networks were designed for the CIFAR-10 and CIFAR-100 datasets, respectively. They share the same architecture described in Table~\ref{table:cifar_baseline_model}, which consists of three sections of layers. The first section is represented by the input layer or ``Stem Convolution". The next section has three stages, each one containing identical building blocks, whose elements are depicted in Figure \ref{fig:block}. This block was designed by introducing the building blocks PyramidNet~\cite{DeepPyramidalResnets:D_Han_2017} in the ResNet-44 architecture~\cite{Resnet:K_He_2016}. The third section consists of a global average-pooling layer followed by a fully-connected layer. Finally, as an important remark, when applying the Entropy-Constrained Trained Ternarization (EC2T) approach, the first and last layers are not quantized. 
\newline
\begin{table*}[h!]
	\renewcommand{\arraystretch}{1.0}
	\caption{Architecture of MicroNet-C10 and MicroNet-C100 networks, where $d$ and $w$ are scaling factors for the networks' depth and width, respectively. For the baseline neworks (i.e., before applying compound-model-scaling), $d=w=1$. The number of classes, $n_{classes}$, corresponds to 10 for CIFAR-10 and 100 for CIFAR-100.} 
	\label{table:cifar_baseline_model}  
	\begin{center}
		\begin{tabular}{ccccc}
			\toprule
			\bfseries Stage & \bfseries Operation & \bfseries Resolution &  \bfseries Output Channels & \bfseries Repetitions\\
			\midrule
			& Stem Convolution ($3\times 3$)   & & &\\
			& $+$ BN $\&$ ReLU & $32\times 32$ & $16\times w$ & 1  \\
			\midrule
			1 & Building Block & $32\times 32$ & $16\times w$ & $7\times d$ \\
			2 & Building Block & $16\times 16$ & $32\times w$ & $7\times d$ \\
			3 & Building Block & $8\times 8$ & $64\times w$ & $7\times d$ \\
			\midrule
			& ReLU $\&$ Global Avg. Pooling  & $8\times 8$ & $64\times w$ & 1 \\
			& Fully-Connected & $1\times~1$ & $n_{classes}$ & 1\\
			\bottomrule
		\end{tabular}
	\end{center} 
\end{table*}
\newline

\newpage
\section{Efficient Storage of Sparse \& Ternary Weight Matrices}
\label{appendix:efficient-parameter-storage}

In addition to the trainable network parameters, we count those values that are needed to reconstruct the model from sparse matrix formats, i.e., binary masks or indices. Specifically, full-precision parameters (32-bits) count as one, while quantized parameters (with less than 32-bits) as a fraction of a parameter. For instance, a binary mask element counts as 1/32 with respect to a full-precision (32-bit) parameter. 

\begin{figure*}[h!]
	\begin{center}
		\includegraphics[width=0.60\textwidth]{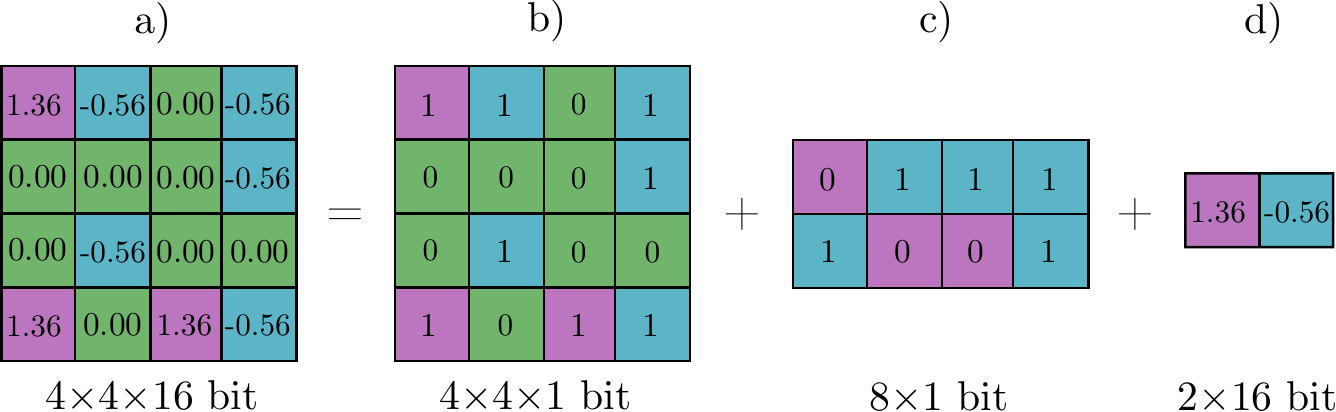}
	\end{center}
	\caption{Efficient storage of sparse and ternary weight matrices. }
	\label{fig:ternarystorage}
\end{figure*}

If Compressed-Entropy-Row(CER)/Compressed-Sparse-Row (CSR) formats are not applied, a ternary convolution layer of size $\mathcal{N}K^2\mathcal{M}$ consists of two binary masks as illustrated in Figure~\ref{fig:ternarystorage}. One mask indicates the location of the centroid values (see Figure~\ref{fig:ternarystorage}b), while the other describes the sign of those values (see Figure~\ref{fig:ternarystorage}c). Thus, the parameter count for these masks is $1/32 \times \mathcal{N}K^2\mathcal{M}$ and $1/32 \times \sigma \mathcal{N}K^2\mathcal{M}$, respectively. In this notation, $\mathcal{N}$ is the number of effective input channels, $K$ the kernel size, $\mathcal{M}$ the number of effective output channels, and $\sigma=1-\text{sparsity}$, with $\sigma\in[0,1]$. The effective number of channels is computed as the original number of channels minus the number of channels pruned by the Entropy-Constrained Trained Ternarization (EC2T) approach. To calculate the layers' sparsity, we exclude the pruned channels. The third matrix in Figure~\ref{fig:ternarystorage}, uses two 16-bit numbers to represent the centroid values. Thus, they count as a single full-precision (32-bit) parameter (Figure \ref{fig:ternarystorage}d). For the batch normalization layers, we add a 16-bit value (bias) per effective output channel. Therefore, their corresponding parameter count is $\mathcal{M}/2$. 
\newline

\end{document}